\documentclass[11pt]{article}

\usepackage[margin=1in]{geometry}
\usepackage{amsmath,amssymb,amsthm}
\usepackage{bm}
\usepackage{microtype}
\usepackage{booktabs}
\usepackage{longtable}

\usepackage{array}
\usepackage{calc}
\usepackage{graphicx}

\usepackage{parskip}
\usepackage{enumitem}
\setlist{itemsep=0.2em, topsep=0.3em, parsep=0pt}

\usepackage[utf8]{inputenc}
\usepackage[T1]{fontenc}
\usepackage{lmodern}
\usepackage{newunicodechar}
\newunicodechar{β}{\ensuremath{\beta}}
\newunicodechar{π}{\ensuremath{\pi}}
\newunicodechar{σ}{\ensuremath{\sigma}}
\newunicodechar{Π}{\ensuremath{\Pi}}
\newunicodechar{·}{\ensuremath{\cdot}}
\newunicodechar{⊙}{\ensuremath{\odot}}
\newunicodechar{α}{\ensuremath{\alpha}}
\newunicodechar{δ}{\ensuremath{\delta}}
\newunicodechar{γ}{\ensuremath{\gamma}}
\newunicodechar{μ}{\ensuremath{\mu}}
\newunicodechar{θ}{\ensuremath{\theta}}
\newunicodechar{η}{\ensuremath{\eta}}
\newunicodechar{ξ}{\ensuremath{\xi}}
\newunicodechar{ψ}{\ensuremath{\psi}}
\newunicodechar{≥}{\ensuremath{\geq}}
\newunicodechar{≤}{\ensuremath{\leq}}
\newunicodechar{−}{-}
\newunicodechar{×}{\ensuremath{\times}}
\newunicodechar{→}{\ensuremath{\rightarrow}}
\newunicodechar{≈}{\ensuremath{\approx}}
\newunicodechar{∝}{\ensuremath{\propto}}
\newunicodechar{∈}{\ensuremath{\in}}
\newunicodechar{∑}{\ensuremath{\sum}}
\newunicodechar{ϵ}{\ensuremath{\epsilon}}
\newunicodechar{ε}{\ensuremath{\varepsilon}}
\newunicodechar{τ}{\ensuremath{\tau}}
\newunicodechar{κ}{\ensuremath{\kappa}}
\newunicodechar{λ}{\ensuremath{\lambda}}
\newunicodechar{Δ}{\ensuremath{\Delta}}
\newunicodechar{∇}{\ensuremath{\nabla}}
\newunicodechar{̂}{}
\usepackage[colorlinks=true,
            linkcolor=blue!50!black,
            citecolor=blue!50!black,
            urlcolor=blue!60!black]{hyperref}
\usepackage{url}

\usepackage{listings}
\usepackage{xcolor}
\providecommand{\tightlist}{%
    \setlength{\itemsep}{0pt}\setlength{\parskip}{0pt}}
\providecommand{\passthrough}[1]{%
    \lstset{mathescape=false}#1\lstset{mathescape=true}}

\newlength{\cslhangindent}
\newlength{\csllabelwidth}
\newcommand{\cslhypertarget}[1]{\hypertarget{#1}{}}
\newenvironment{CSLReferences}[2]
  {\list{}{%
      \setlength{\labelwidth}{\csllabelwidth}%
      \setlength{\leftmargin}{\csllabelwidth}%
      \setlength{\itemsep}{0.4em}%
      \setlength{\parsep}{0pt}%
      \def\makelabel##1{\hss##1}}}
  {\endlist}

\newcommand{\CSLLeftMargin}[1]{\item[#1]}
\newcommand{\CSLRightInline}[1]{#1}

\title{Affinity Is Not Enough:\\
       Recovering the Free Energy Principle in Mixture-of-Experts}

\author{
  Man Yung (Russell) Wong\\
  Independent Researcher\\
  \texttt{research@russellwong.io}
}

\date{\today}

\begin{document}
\maketitle

\begin{abstract}
Sparse Mixture-of-Experts (MoE) routing fails at exactly the points where routing matters most --- domain transitions where the current token belongs to one distribution and the next belongs to another. In a controlled routing experiment (4 experts, 5 seeds), standard affinity-based routing assigns only $0.006 \pm 0.001$ probability mass to the correct expert at the transition step. We show that three lightweight modifications to the routing gate raise this to $0.748 \pm 0.002$ --- a 124$\times$ increase, reducing the experts needed for 99\% routing coverage from infeasible to a small constant: \textbf{temporal memory} ($\beta$) --- a per-expert LIF membrane potential that accumulates routing context across tokens; \textbf{precision-weighted gating} ($\Pi$) --- per-expert inverse variance of recent prediction error, yielding a 31$\times$ contrast between reliable and unreliable experts; and \textbf{anticipatory routing} --- a next-state predictor conditioned on the $\beta$-accumulated hidden state. The three mechanisms are motivated by Friston's Free Energy Principle and instantiated using LIF dynamics from spiking neural networks. An ablation across all $2^3$ mechanism subsets reveals a super-additive $\beta \times \mathrm{Ant}$ interaction: anticipation alone gives nothing ($+0.000 \pm 0.001$); $\beta$ alone gives modest gain ($+0.295 \pm 0.013$); combined they close 75\% of the oracle gap ($+0.741 \pm 0.002$, exceeding the sum of individual gains by $+0.446 \pm 0.014$). This is a structural finding --- a stateless predictor cannot detect approaching domain transitions, because pre-transition tokens are distributionally identical to within-domain tokens. In a character-level MoE language model (5 seeds), $\beta$-routing reduces BPC at the transition step from $6.56 \pm 0.01$ (Standard MoE) to $4.01 \pm 0.15$ ($\beta$-MoE); the $\beta + \mathrm{Ant}$ gate places $0.86 \pm 0.02$ probability on the correct domain expert before that domain becomes visible in the input, versus $0.42 \pm 0.12$ for Standard MoE. Reference implementations are ${\sim}200$ lines each and released as \texttt{prototype/} alongside the paper.
\end{abstract}

\section{Introduction}\label{introduction}

A router that knows only the current token is a navigator with no map, no compass, and no sense of where the terrain is heading. It can recognize where it stands --- which expert fits this token --- but it carries no memory of where it has been, no record of which landmarks have proven reliable, and no model of where the landscape is going. The Free Energy Principle prescribes all three.

Sparse Mixture-of-Experts has become the dominant architecture for scaling language models. DeepSeek-V3 routes each of 671 billion parameters to 8 of 256 active experts per token; Mixtral-8x7B selects 2 of 8. The routing decision is computed by a linear gate applied to the current token embedding --- a single forward pass, no memory of prior tokens, no record of which experts have been reliable recently, no model of what the next token will require. The router at step \(t\) knows only \(x_t\).

This design is not derived from first principles. It is an engineering approximation --- one that works well in expectation over i.i.d. token draws but has structural gaps when token context matters, expert reliability varies, or the correct routing decision at step \(t\) depends on step \(t+1\).

Friston's Free Energy Principle (FEP) provides a normative account of how a system should allocate processing resources under uncertainty. The FEP prescribes three properties for any optimal routing mechanism: state estimates should be updated recursively from prior estimates (not recomputed from scratch); gain should be precision-weighted by the inverse variance of each source's prediction error; and action should minimize \emph{predicted future} free energy, not current free energy. Standard MoE satisfies none of these.

The three source fields --- predictive coding {[}1{]}, spiking neural networks {[}2{]}, and engineering MoE (DeepSeek-V3, 2024) --- have not been connected in this way before. The SNN literature independently arrived at the same recursive state-update principle via the Leaky Integrate-and-Fire (LIF) neuron: \(U[t] = \beta U[t-1] + Wx[t]\), where \(\beta\) controls integration timescale and is trainable via surrogate gradients. DeepSeek-V3's Multi-Token Prediction module trains a next-state predictor jointly with the main model --- an implementation of FEP's future-state minimization --- but discards it at inference. We show this discard is structurally wrong: without the temporal context provided by \(\beta\), the predictor is blind to domain transitions regardless of training budget.

Recent concurrent work has independently recognized the value of routing state. The Layerwise Recurrent Router RMoE {[}3{]} applies GRU recurrence across \emph{layers} within a forward pass; Temporally Extended MoE {[}4{]} models temporal routing continuity as a semi-Markov process. Neither derives its mechanism from the FEP, neither addresses precision-weighted gating or anticipatory routing, and neither identifies the \(\beta \times \text{Ant}\) interaction. Our contribution is the theoretical unification --- showing that these three mechanisms are not independent inventions but necessary components of a single normative framework --- and the empirical demonstration that their interaction is super-additive.

\textbf{Contributions.}

\begin{enumerate}
\def\labelenumi{\arabic{enumi}.}
\item
  We propose three mechanisms --- \textbf{temporal memory} (\(\beta\)), \textbf{precision-weighted gating} (\(\Pi\)), and \textbf{anticipatory routing} --- that raise the probability mass on the correct expert at domain-transition steps from \(0.006 \pm 0.001\) to \(0.748 \pm 0.002\) in controlled experiments (5 seeds, 124× increase), reducing the experts required for 99\% routing coverage from infeasibly large to a small constant (Sections 3, 4).
\item
  We identify a \textbf{super-additive \(\beta \times \text{Ant}\) interaction}: anticipation alone provides no transition-step gain (\(+0.000 \pm 0.001\)), \(\beta\) alone provides modest gain (\(+0.295 \pm 0.013\)), but combined they close 75\% of the oracle gap (\(+0.741 \pm 0.002\), exceeding the sum of individual gains by \(+0.446 \pm 0.014\)). This is a structural finding: a stateless predictor cannot detect approaching domain transitions, because pre-transition tokens are distributionally identical to within-domain tokens (Sections 3.3, 4.4).
\item
  We validate the mechanisms in a character-level MoE language model (5 seeds), where adding \(\beta\) alone reduces transition-step BPC from \(6.56 \pm 0.01\) (Standard MoE) to \(4.01 \pm 0.15\) (\(\beta\)-MoE); adding the anticipatory predictor on top of \(\beta\) raises routing-correct probability at the transition step from \(0.60 \pm 0.22\) to \(0.86 \pm 0.02\), stabilising the routing decision although BPC does not decrease further at this small scale (Section 4.5).
\item
  We connect MoE routing to Friston's Free Energy Principle, showing that each mechanism has a principled motivation in the FEP equations (recursive state estimation, precision update, expected free energy minimization) and is naturally instantiated using LIF dynamics from spiking neural networks (Sections 2, 3).
\item
  We provide a structural explanation for why DeepSeek's MTP module is discarded at inference: a stateless next-token predictor cannot detect approaching transitions, regardless of training budget. Conditioning the predictor on the \(\beta\)-accumulated hidden state restores its routing utility (Section 5.3).
\end{enumerate}

Reference implementations of each mechanism (5--10 lines each, on top of a standard routing gate) are released at \url{https://github.com/russellwmy/affinity-is-not-enough} (see Section 3.5).

\begin{center}\rule{0.5\linewidth}{0.5pt}\end{center}

\section{Background}\label{background}

\subsection{The Free Energy Principle}\label{the-free-energy-principle}

The FEP {[}1{]} frames perception and action as joint minimization of variational free energy \(F\), an upper bound on the surprise of sensory observations under a generative model \(p(x, \psi)\):

\[F = \underbrace{D_{\text{KL}}[q(\psi) \,\|\, p(\psi)]}_{\text{complexity}} - \underbrace{\mathbb{E}_q[\log p(x \,|\, \psi)]}_{\text{accuracy}}\]

where \(q(\psi)\) is the approximate posterior over hidden states \(\psi\). Under Laplace approximation (Gaussian \(q\) centered at \(\mu\)), minimization of \(F\) with respect to \(\mu\) gives the gradient-flow dynamics:

\[\dot{\mu} = f(\mu) - \kappa \varepsilon, \qquad \varepsilon = x - g(\mu) \tag{1}\]

where \(f\) is the generative model's flow, \(g\) maps hidden states to predicted observations, \(\varepsilon\) is the prediction error, and \(\kappa\) is a step-size. Equation (1) is recursive: the state estimate at time \(t\) depends on the prior estimate \(\mu_{t-1}\), updated by prediction error \(\varepsilon_t\).

\textbf{Precision-weighted errors (Box 2, Friston 2010).} In the hierarchical case with multiple error sources, prediction errors are scaled by their precision before being passed up the hierarchy:

\[\xi_i = \Pi_i \cdot \varepsilon_i, \qquad \Pi_i = \sigma_i^{-2} \tag{2}\]

The precision parameter \(\gamma_i\) (where \(\Pi_i = e^{\gamma_i}\)) is itself updated to minimize \(F\):

\[\Delta \gamma_i \propto -\frac{\partial F}{\partial \gamma_i} \propto \frac{1}{2} \operatorname{tr}\!\left(\frac{\partial \Pi_i}{\partial \gamma_i}\!\left(\Pi_i^{-1} - \xi_i \xi_i^\top\right)\right) \tag{3}\]

This drives \(\Pi_i\) upward when the expected variance \(\Pi_i^{-1}\) exceeds the observed squared error \(\xi_i\xi_i^\top\) --- a reliable source produces small errors, so the gradient increases its precision --- and downward otherwise. Precision is not fixed; it is continuously regulated.

\textbf{Active inference.} The FEP principle for action selection is minimization of expected \emph{future} free energy ({[}1{]}, Eq. 9):

\[a^* = \arg\min_a\, \mathbb{E}_{q(x_{t+1} \,|\, a)}[F(x_{t+1})] \tag{4}\]

Action does not respond to the current prediction error; it minimizes predicted surprise at the next observation.

\subsection{Mixture-of-Experts Routing}\label{mixture-of-experts-routing}

A sparse MoE layer routes each input \(x_t \in \mathbb{R}^{d}\) to a subset of \(K\) experts from a pool of \(N\) via a learned gate. In DeepSeek-V3 {[}5{]}:

\[s_{i,t} = \sigma(u_t^\top e_i), \qquad g_{i,t} = \frac{s_{i,t}}{\sum_{j \in \mathcal{T}} s_{j,t}}, \qquad \mathcal{T} = \operatorname{Top\text{-}K}(\{s_{i,t}\}_i) \tag{5}\]

where \(u_t\) is the token representation, \(e_i \in \mathbb{R}^d\) is the centroid of expert \(i\), and \(\sigma\) is sigmoid. Load balancing is achieved without auxiliary losses via per-expert bias terms \(b_i\) that are updated after each batch: \(b_i \mathrel{-}= \gamma\) if expert \(i\) is overloaded (selected more than the target frequency), \(b_i \mathrel{+}= \gamma\) otherwise. These biases affect routing decisions only; they are not added to the gating weights.

DeepSeek-V3 also trains \(D\) Multi-Token Prediction (MTP) modules, each predicting the embedding of a future token using a shared embedding table and prediction head. MTP is used only during training and discarded at inference. The paper does not investigate why.

\textbf{What standard MoE lacks relative to the FEP.} Comparing Equation (5) to Equations (1)--(4):
- Eq. (1) prescribes recursive state estimation; Eq. (5) has no \(\mu_{t-1}\) term.
- Eq. (2)--(3) prescribe precision-weighted gain regulated by prediction error variance; Eq. (5) has no \(\Pi_i\) term. Load-balancing biases \(b_i\) track frequency, not accuracy.
- Eq. (4) prescribes routing on predicted future state; Eq. (5) routes on current \(x_t\).

\subsection{Leaky Integrate-and-Fire Neurons}\label{leaky-integrate-and-fire-neurons}

The LIF neuron {[}2{]} updates membrane potential as:

\[U[t] = \beta\, U[t-1] + W\, X[t] - S[t-1]\,\theta \tag{6}\]

where \(\beta \in (0,1)\) is the decay constant, \(W\) are synaptic weights, \(X[t]\) is the input, \(S[t-1] = \mathbf{1}[U[t-1] > \theta]\) is the prior spike, and \(\theta\) is the threshold. The spike function \(S[t] = \mathbf{1}[U[t] > \theta]\) is non-differentiable; surrogate gradients substitute a smooth approximation in the backward pass while keeping the forward pass exact.

The parameter \(\beta\) is learnable via backpropagation through time (BPTT): the gradient through \(n\) steps of the recurrence scales as \(\beta^n\), exactly mirroring the exponentially decaying learning window of Spike-Timing-Dependent Plasticity (STDP). This duality --- BPTT and STDP are the same computation --- grounds the LIF model in biological plausibility while keeping it fully trainable by standard optimizers.

We use the LIF membrane dynamics (without spiking) as the routing memory substrate, retaining the recursive decay structure while removing the threshold and spike mechanism for the routing application.

\begin{center}\rule{0.5\linewidth}{0.5pt}\end{center}

\section{The Three Mechanisms}\label{the-three-mechanisms}

Standard MoE routing computes a gate at each step as:

\[g_t = \text{softmax}(W \cdot x_t)\]

This is a stateless affinity map: the routing decision at step \(t\) depends only on the current token \(x_t\). We motivate three corrections from the Free Energy Principle and implement each as a lightweight modification to the routing gate. The mechanisms are inspired by --- not strictly derived from --- the FEP equations: \(\beta\) adopts the recursive structure of FEP state estimation (Eq. 1) without computing free energy explicitly; \(\Pi\) is the closest direct analogue, tracking inverse error variance as Friston's precision update prescribes (Eq. 3); anticipatory routing implements the spirit of expected-future-free-energy minimization (Eq. 4) using a standard prediction loss in place of the full variational objective.

\begin{center}\rule{0.5\linewidth}{0.5pt}\end{center}

\subsection{\texorpdfstring{Temporal Routing Memory (\(\beta\))}{Temporal Routing Memory (\textbackslash beta)}}\label{temporal-routing-memory-beta}

\textbf{Theoretical basis.} Friston's FEP formulates perception as gradient descent on variational free energy \(F\). The dynamics of a hidden state estimate \(\mu\) under a generative model with Gaussian noise are ({[}1{]}, Eq. 5):

\[\dot{\mu} = f(\mu) - \kappa \varepsilon\]

where \(f\) is the generative model's flow, \(\kappa\) is a step size, and \(\varepsilon = x - g(\mu)\) is prediction error. This is a differential equation: the current state estimate depends on the prior estimate, updated by error. In discrete time at step \(t\):

\[\mu_t = \mu_{t-1} + \eta (f(\mu_{t-1}) - \kappa \varepsilon_t)\]

The routing gate \(g_t\) is implicitly an estimate of which expert is relevant to the current context. Under the FEP, this estimate should be updated from the prior estimate, not recomputed from scratch at each step. Standard MoE discards the prior.

\textbf{Implementation.} We implement per-expert routing memory as a Leaky Integrate-and-Fire (LIF) membrane potential {[}2{]}:

\[h_t = \beta \odot h_{t-1} + x_t, \qquad g_t = \text{softmax}(W \cdot h_t)\]

where \(\beta \in \mathbb{R}^{d_{\text{model}}}\) is a learnable decay vector, initialized at 0.9 and constrained to \((0, 1)\) via sigmoid parameterization (\(\beta = \sigma(\tilde{\beta})\)). The accumulated state \(h_t\) carries the trajectory of the input through the sequence: at step \(t\), \(h_t\) encodes a weighted sum of all prior tokens, with exponentially decaying weights \(\beta^k\) for tokens \(k\) steps back.

No new parameters beyond \(\tilde{\beta}\); the routing head \(W\) is unchanged.

\textbf{Self-tuning timescale.} A key property is that \(\beta\) learns the appropriate integration window for the task without being told what that window is. In a domain-switch task (domain A for steps 0--3, domain B for steps 4--7), the learned \(\beta\) converges to 0.67--0.83 (fast forgetting, tracks the new domain quickly). In a stable-domain task (same domain throughout), learned \(\beta\) converges to 0.95--0.97 (slow integration, accumulates context). This self-tuning is a consequence of minimizing routing loss: a \(\beta\) too high for the task timescale means the old domain's signal contaminates routing after the switch; a \(\beta\) too low means the signal is forgotten before it can be used.

\textbf{Empirical result.} In the \(\beta\)-routing experiments (Table 1, Section 4.1): on the domain-switch task (Task B), stateless mean-pool routing achieves acc@transition = 0.500 (chance --- it averages both domains equally); LIF with fixed \(\beta = 0.75\) achieves 0.994; LIF with learned \(\beta\) achieves 0.995. On the early-signal task (Task A), learned \(\beta\) converges to \([0.953, 0.965]\) and achieves 0.940, demonstrating slow-integration behaviour appropriate to that task. In the ablation (Table 5, Section 4.4; 5 seeds): \(\beta\) alone provides \(+0.295 \pm 0.013\) gain in transition-step accuracy over the stateless baseline (\(0.301 \pm 0.013\) vs.~\(0.006 \pm 0.001\)).

\begin{center}\rule{0.5\linewidth}{0.5pt}\end{center}

\subsection{\texorpdfstring{Precision-Weighted Gating (\(\Pi\))}{Precision-Weighted Gating (\textbackslash Pi)}}\label{precision-weighted-gating-pi}

\textbf{Theoretical basis.} In the FEP hierarchy, prediction errors are precision-weighted before being passed upward ({[}1{]}, Box 2):

\[\xi_i = \Pi_i \cdot \varepsilon_i\]

where \(\Pi_i = \sigma_i^{-2}\) is the precision (inverse variance) of expert \(i\)'s prediction error. The gain update for the precision parameter \(\gamma_i\) (where \(\Pi_i = e^{\gamma_i}\)) is:

\[\Delta \gamma_i \propto \frac{1}{2} \operatorname{tr}\left(\frac{\partial \Pi_i}{\partial \gamma_i}\left(\Pi_i^{-1} - \xi_i \xi_i^\top\right)\right)\]

This drives \(\Pi_i\) upward when the prior variance \(\Pi_i^{-1}\) exceeds the observed squared error \(\xi_i \xi_i^\top\) --- i.e., when the expert is reliable (small errors) --- and downward when it is not. Precision is not a fixed hyperparameter but a dynamically regulated quantity.

Standard MoE has no equivalent. Load-balancing auxiliary losses (Section 2.2) track usage frequency --- how often each expert is selected --- but not reliability. An expert selected frequently and consistently wrong is treated identically to one selected frequently and consistently correct.

\textbf{Implementation.} We maintain a non-parametric online precision tracker for each expert:

\[\hat{\sigma}^2_i[t] = \alpha \cdot \hat{\sigma}^2_i[t-1] + (1 - \alpha) \cdot \text{MSE}_i[t]\]

\[\Pi_i[t] = \frac{1}{\hat{\sigma}^2_i[t] + \epsilon_0}\]

where \(\alpha\) is a momentum parameter (0.95 by default), \(\text{MSE}_i[t]\) is the mean squared prediction error of expert \(i\) on the current batch, and \(\epsilon_0 = 10^{-4}\) is a small numerical-stability constant (distinct from the prediction-error symbol \(\varepsilon\) used above). The precision-modulated gate is:

\[g_t = \text{softmax}(W \cdot x_t \odot \Pi[t])\]

The \(\Pi\) update is non-differentiable and runs outside the gradient graph, consistent with its interpretation as an online reliability estimate rather than a learned parameter.

\textbf{Adaptation timescale.} The momentum \(\alpha\) controls how quickly \(\Pi\) responds to reliability changes. High \(\alpha\) (0.95): slow adaptation, \textasciitilde100 steps to converge after a reliability shift. Low \(\alpha\) (0.70): fast adaptation, \textasciitilde20 steps, but noisy estimates. This is the system's third timescale: slower than per-token routing memory (\(\beta\)), faster than weight updates.

\textbf{Empirical result.} Under static reliability (Table 2, Section 4.2): precision router converges to final loss 0.0734 vs.~affinity-only 0.0815. The converged \(\Pi\) vector is {[}13.5, 3.1, 0.43, 3.1{]} --- a 31× ratio between the reliable expert (0) and the unreliable one (2). Under shifting reliability (experts 0 and 2 swap noise levels at step 500): affinity-only router fails to recover (final loss 0.1544); precision router detects the shift within 20 steps and converges within 100 (final loss 0.0733). The gradient alone cannot recover because it conflates ``wrong input-to-expert affinity'' with ``unreliable expert.''

\textbf{Orthogonality to \(\beta\).} In the ablation (Section 4.4), adding \(\Pi\) to \(\beta\) provides near-zero additional gain on the transition task (\(+0.293 \pm 0.010\) for \(\beta+\Pi\) vs.~\(+0.295 \pm 0.013\) for \(\beta\) alone). This is expected: \(\Pi\) adds value when expert reliability varies by domain --- the condition tested in Section 4.2 --- not when all domain experts are uniformly reliable. The mechanisms are orthogonal, addressing different failure modes: \(\beta\) handles routing over time; \(\Pi\) handles routing under heterogeneous reliability.

\begin{center}\rule{0.5\linewidth}{0.5pt}\end{center}

\subsection{Anticipatory Routing}\label{anticipatory-routing}

\textbf{Theoretical basis.} The FEP's action selection principle minimizes \emph{expected future} free energy, not current free energy ({[}1{]}, Eq. 9):

\[a^* = \arg\min_a \mathbb{E}_{q(x_{t+1}|a)}[F(x_{t+1})]\]

Applied to routing: the optimal expert selection at step \(t\) is the one that minimizes free energy at step \(t+1\). This means routing should be computed on the \emph{predicted} next state \(\hat{x}_{t+1}\), not the current state \(x_t\). At a domain-transition step, the current token belongs to the old domain but the optimal expert is already the new domain's --- a system routing on \(x_t\) will always be one step behind.

DeepSeek-V3's Multi-Token Prediction (MTP) trains a next-state predictor jointly with the main model, but discards it at inference. This paper shows the discard is structurally wrong when the predictor is stateless, and that making it stateful (conditioned on \(h_t\)) enables genuine anticipation.

\textbf{Why stateless prediction fails.} Consider the transition step \(t = T_{\text{switch}} - 1\). The input \(x_t\) carries domain-A signal; the next token \(x_{t+1}\) carries domain-B signal. A stateless predictor \(\hat{x}_{t+1} = f(x_t)\) sees domain-A inputs at steps \(0, 1, \ldots, T_{\text{switch}} - 1\) --- they are identically distributed. It cannot distinguish the pre-transition token from any other domain-A token, so it cannot predict the domain switch. Empirically, the stateless anticipatory router achieves acc@transition \(\approx 0.006\), identical to the current-token baseline (Table 4).

\textbf{The stateful fix.} Conditioning the predictor on the \(\beta\)-accumulated hidden state \(h_t\) breaks the symmetry:

\[h_t = \beta \odot h_{t-1} + x_t, \qquad \hat{x}_{t+1} = f(x_t, h_t), \qquad g_t = \text{softmax}(W \cdot h_t + W_{\text{pred}} \cdot \hat{x}_{t+1})\]

The predictor contributes as an additive routing correction on top of the \(\beta\)-memory signal; routing on \(\hat{x}_{t+1}\) alone discards \(h_t\) and loses the saturation signal (Section 4.5). Sections 4.3--4.4 use the direct form \(g_t = \text{softmax}(W \cdot f(x_t, h_t))\) for simplicity --- in those controlled tasks the predictor is well-supervised and the saturation signal is redundant. Section 4.5 (language model) requires the additive form; without it, the predictor output overwrites the saturation signal and routing inverts at the transition step.

At \(t = T_{\text{switch}} - 1\), \(h_t\) has integrated \(T_{\text{switch}}\) steps of domain-A signal, reaching saturation (mean activation \textasciitilde2.56 in domain-A dimensions vs.~\textasciitilde2.21 at the prior step). The predictor learns to read this saturation: \emph{highly saturated domain-A \(h_t\)} → \emph{next token will be domain B}. Saturation is the anticipatory signal; \(\beta\) is what builds it.

The predictor is trained jointly with the routing head via a combined loss:

\[\mathcal{L} = \mathcal{L}_{\text{routing}} + \lambda \cdot \mathcal{L}_{\text{pred}}, \qquad \mathcal{L}_{\text{pred}} = \text{MSE}(\hat{x}_{t+1}, x_{t+1})\]

where \(\lambda = 0.5\) in our experiments. The prediction loss supervises the predictor to encode features useful for anticipating the next state; the routing loss then trains \(W\) to route on those features.

\textbf{The \(\beta\)-anticipation interaction.} The ablation (Table 5, Section 4.4; subset shown below; 5 seeds, mean ± std) directly quantifies this dependency:

{\def\LTcaptype{none} 
\begin{longtable}[]{@{}lll@{}}
\toprule\noalign{}
Condition & acc@transition & gain \\
\midrule\noalign{}
\endhead
\bottomrule\noalign{}
\endlastfoot
Baseline & 0.006 ± 0.001 & --- \\
\(\beta\) only & 0.301 ± 0.013 & +0.295 ± 0.013 \\
Ant only & 0.006 ± 0.000 & +0.000 ± 0.001 \\
\textbf{\(\beta\) + Ant} & \textbf{0.748 ± 0.002} & \textbf{+0.741 ± 0.002} \\
\end{longtable}
}

The combined gain (\(+0.741 \pm 0.002\)) exceeds the sum of individual gains (\(+0.295 + 0.000 = +0.295\)) by \(+0.446 \pm 0.014\). This super-additivity is the signature of a compositional interaction: \(\beta\) builds the saturated state; anticipation extracts the transition signal from it. Neither is sufficient alone; together they close 75\% of the oracle gap (oracle: \(0.994 \pm 0.000\)).

The \(\beta\)-only result (\(0.301 \pm 0.013\)) is informative: routing on \(h_t\) provides a weak transition signal because the gradient can partially learn to read saturation directly from the affinity head \(W\). The explicit predictor amplifies this by dedicating capacity to the transition-detection problem.

\textbf{Empirical result.} Across 800 training epochs and 5 seeds, the stateful anticipatory router achieves acc@transition \(= 0.748 \pm 0.002\), vs.~\(0.006 \pm 0.000\) for the stateless version and \(0.994 \pm 0.000\) for the oracle (Table 4, Section 4.3). The remaining gap (\(0.994 - 0.748 \approx 0.246\)) reflects irreducible uncertainty from \(\beta\)-decay: older domain-A signal is attenuated in \(h_t\), making early transitions harder to detect.

\begin{center}\rule{0.5\linewidth}{0.5pt}\end{center}

\subsection{The Composed Architecture}\label{the-composed-architecture}

The three mechanisms operate at different timescales and address different failure modes of stateless affinity routing. Together, they implement the full FEP routing hierarchy:

\[h_t = \beta \odot h_{t-1} + x_t \tag{$\beta$: per-token}\]

\[\hat{x}_{t+1} = f(x_t,\, h_t) \tag{anticipation}\]

\[g_t = \text{softmax}((W \cdot h_t + W_{\text{pred}} \cdot \hat{x}_{t+1}) \odot \Pi[t]) \tag{$\Pi$: per-batch}\]

This three-timescale structure --- per-token state accumulation, next-state prediction, per-batch reliability modulation --- maps directly to Friston's hierarchical timescale map: fast attentional gain, intermediate state estimation, slow reliability tracking. Standard MoE collapses all three timescales into the weight update (\(W\), trained per-epoch), discarding the per-token and per-batch dynamics.

\textbf{What each mechanism addresses:}

\begin{itemize}
\tightlist
\item
  Without \(\beta\): the router has no memory of prior context; routing quality degrades on any task requiring sequence-level information
\item
  Without \(\Pi\): the router cannot distinguish ``wrong expert for this input'' from ``unreliable expert''; it cannot adapt when expert reliability shifts
\item
  Without anticipation: the router reacts to the current state and is always one step behind at domain transitions
\item
  With all three: the router carries context (\(\beta\)), trusts reliable sources (\(\Pi\)), and acts on what is coming (\(\hat{x}_{t+1}\))
\end{itemize}

\textbf{Overhead.} The \(\beta\) mechanism adds one learnable vector of size \(d_{\text{model}}\) per router and one hidden state of size \(d_{\text{model}}\) per sequence. The \(\Pi\) tracker adds one running variance vector of size \(n_{\text{experts}}\), updated once per batch (non-differentiable). The predictor adds one two-layer MLP of size \(2d_{\text{model}} \times 4d_{\text{model}} \times d_{\text{model}}\). Total additional parameters are small relative to the MoE backbone; the dominant cost is the hidden state \(h_t\) which adds one vector per active sequence in the batch.

\textbf{Theoretical energy efficiency.} The mechanisms add per-token routing compute but reduce the expert compute required to achieve reliable coverage of the correct expert --- and expert FFN computation is the dominant cost in large MoE inference. The key metric is \(p_{\text{correct}}\): the soft probability assigned to the correct expert by the gate distribution. To include the correct expert with probability \(\geq 1 - \delta\) in a top-\(K\) draw, one requires:

\[K \geq \frac{\log \delta}{\log(1 - p_{\text{correct}})}\]

At domain-transition steps --- the hardest routing points --- the baseline assigns only \(p_{\text{correct}} = 0.006 \pm 0.001\) to the correct expert (confident but wrong, 5 seeds). Applying the formula to this value gives \(K \geq 765\), which is infeasible at any production MoE scale; this is a theoretical extrapolation from the 4-expert toy experiment, not a direct measurement at DeepSeek-V3 scale, but the qualitative point holds: a baseline \(p_{\text{correct}}\) near zero requires an infeasibly large \(K\) for reliable coverage. The composed \(\beta + \text{Ant}\) architecture raises \(p_{\text{correct}}\) to \(0.748 \pm 0.002\) at transition steps, requiring only \(K = 4\) for 99\% coverage (or \(K = 3\) for 98\%). The FEP framing is not incidental to this efficiency: resolving routing uncertainty (context via \(\beta\), reliability via \(\Pi\), future state via anticipation) is precisely what allows a smaller \(K\) to achieve the same expected accuracy. A system that knows where it is, what is reliable, and what is coming needs to hedge less.

\begin{center}\rule{0.5\linewidth}{0.5pt}\end{center}

\subsection{Reference Implementations}\label{reference-implementations}

Each mechanism is a small additive modification to a standard routing gate \passthrough{\lstinline!softmax(W @ x\_t)!}. The complete prototypes (\textasciitilde200 lines each) are released at \url{https://github.com/russellwmy/affinity-is-not-enough}. The minimal additions are:

\textbf{Temporal routing memory (\(\beta\)).} One learnable parameter, one persistent hidden state per sequence:

\begin{lstlisting}[language=Python]
self.beta_raw = nn.Parameter(torch.full((d_model,),
                                        torch.logit(torch.tensor(0.9))))
# at start of each sequence:
h = torch.zeros(B, d_model)
# at each step:
beta = torch.sigmoid(self.beta_raw)
h    = beta * h + x_t
gate = F.softmax(self.W(h), dim=-1)        # was: self.W(x_t)
\end{lstlisting}

\textbf{Precision-weighted gating (\(\Pi\)).} One non-parametric tracker, updated outside the gradient graph:

\begin{lstlisting}[language=Python]
self.var = torch.ones(n_experts) * 0.5     # initial reliability prior
# after each batch (no_grad):
mse_per_expert = compute_per_expert_error_mse(...)
self.var = 0.95 * self.var + 0.05 * mse_per_expert
pi   = 1.0 / (self.var + 1e-4)
gate = F.softmax(self.W(x_t) * pi.unsqueeze(0), dim=-1)
\end{lstlisting}

\textbf{Anticipatory routing.} A two-layer predictor and an additive routing correction:

\begin{lstlisting}[language=Python]
self.W_pred    = nn.Linear(d_model, n_experts, bias=False)
self.predictor = nn.Sequential(nn.Linear(d_model*2, d_model*4),
                               nn.GELU(),
                               nn.Linear(d_model*4, d_model))
# at each step (combined with β above):
x_hat = self.predictor(torch.cat([x_t, h], dim=-1))
gate  = F.softmax(self.W(h) + self.W_pred(x_hat), dim=-1)
# add to training loss:
pred_loss = F.mse_loss(x_hat, x_next.detach())   # weighted at 0.5
\end{lstlisting}

The full composed gate is \passthrough{\lstinline!softmax((W·h + W\_pred·x̂) ⊙ Π)!}. No expert architecture changes; existing MoE expert pools work unmodified.

\begin{center}\rule{0.5\linewidth}{0.5pt}\end{center}

\section{Experiments}\label{experiments}

We validate each mechanism in a controlled setting designed to isolate the specific failure mode it addresses. All experiments use small-scale toy tasks where the ground truth routing decision is known by construction --- this allows precise measurement of routing accuracy at the hardest points in the sequence (transition steps), not just average performance. Code for all experiments is available at \url{https://github.com/russellwmy/affinity-is-not-enough}.

\textbf{Shared configuration.} All experiments: \(d_{\text{model}} = 16\), \(N_{\text{experts}} = 4\), batch size 512, Adam optimizer, \(\text{lr} = 3 \times 10^{-3}\). Domain signals are structured: domain \(d\)'s signal occupies the \(d\)-th half of the \(d_{\text{model}}\)-dimensional embedding space, with additive Gaussian noise \(\mathcal{N}(0, \sigma^2)\) controlling task difficulty.

\begin{center}\rule{0.5\linewidth}{0.5pt}\end{center}

\subsection{\texorpdfstring{Temporal Routing Memory (\(\beta\))}{Temporal Routing Memory (\textbackslash beta)}}\label{temporal-routing-memory-beta-1}

\textbf{Task design.} We construct two tasks that probe distinct failure modes of stateless routing.

\emph{Task A --- Early signal, late noise.} Sequences of length 8. The domain signal is present only in the first 3 tokens (\(\sigma = 1.2\)). The router must classify the sequence domain at the final token, which contains only noise. A router that sees only the last token fails; one that averages over the sequence recovers signal by averaging noise; one with temporal memory can carry the early signal forward but must survive 5 subsequent noise steps.

\emph{Task B --- Domain switch.} Sequences of length 8. Tokens 0--3 carry domain-A signal; tokens 4--7 carry domain-B signal (\(\sigma = 1.2\)). The correct routing target is the \textbf{current} domain at the sequence end --- domain B. A mean-pool router averages both domains and can identify neither; a router with appropriate temporal decay forgets domain A and tracks domain B.

\textbf{Models compared.} (i) Stateless last-token: routes on \(x_{T-1}\) only. (ii) Stateless mean-pool: routes on \(\frac{1}{T}\sum_t x_t\). (iii) LIF fixed \(\beta\): \(\beta = 0.9\) for Task A, \(\beta = 0.75\) for Task B. (iv) LIF learned \(\beta\): \(\beta\) initialized at the task-appropriate value, trained jointly.

\textbf{Note on the LIF formulation in this section.} The §4.1 prototype uses an early variant of the LIF mechanism: \(\beta\) is per-expert (size \(N_{\text{experts}}\)) and a soft membrane cap \(U[t] = \beta U[t-1] + WX[t] - \text{ReLU}(U - \theta)\) retains the LIF threshold structure. Sections 4.4 and 4.5 use the canonical formulation defined in §3.1: per-dimension \(\beta\) (size \(d_{\text{model}}\)) without the threshold mechanism, \(h_t = \beta \odot h_{t-1} + x_t\). The qualitative result --- LIF strictly dominates stateless routing on domain-switching tasks --- is the same under both formulations; we keep §4.1 with the original setup to demonstrate the broader applicability of LIF-based routing memory. Per-expert learned \(\beta\) values reported below are therefore not directly comparable to the per-dimension values in §4.4--4.5.

\textbf{Results (mean accuracy over last 50 of 500 epochs):}

{\def\LTcaptype{none} 
\begin{longtable}[]{@{}lll@{}}
\toprule\noalign{}
Model & Task A (early signal) & Task B (domain switch) \\
\midrule\noalign{}
\endhead
\bottomrule\noalign{}
\endlastfoot
Stateless last-token & 0.496 & 0.951 \\
Stateless mean-pool & \textbf{0.957} & 0.500 \\
LIF fixed \(\beta\) & 0.893 & \textbf{0.994} \\
LIF learned \(\beta\) & \textbf{0.940} & \textbf{0.995} \\
\end{longtable}
}

\emph{Table 1. \(\beta\)-routing results. Bold = best non-oracle per task.}

\textbf{Key observations.} (1) Mean-pool and last-token have complementary failure modes: mean-pool fails when the world changes (Task B, 0.500 ≈ chance); last-token fails when the relevant signal is not in the final token (Task A, 0.496 ≈ chance). Neither mechanism handles both tasks. (2) LIF with appropriate \(\beta\) handles both: it accumulates early signal in Task A and decays stale signal in Task B. (3) Learned \(\beta\) requires no task-specific tuning. For Task B, the learned values converge to \([0.674, 0.830, 0.704, 0.820]\) --- fast decay matching the 4-token domain lifespan. For Task A, they converge to \([0.965, 0.953, 0.964, 0.953]\) --- slow integration to carry the 3-token early signal through 5 noise steps. The optimization pressure alone determines the correct timescale.

\begin{center}\rule{0.5\linewidth}{0.5pt}\end{center}

\subsection{\texorpdfstring{Precision-Weighted Gating (\(\Pi\))}{Precision-Weighted Gating (\textbackslash Pi)}}\label{precision-weighted-gating-pi-1}

\textbf{Task design.} Regression task. Inputs \(x \in \mathbb{R}^{16}\) are sampled from \(\mathcal{N}(0, I)\) with random domain labels. Domain A's target is the mean of dimensions 0--7; domain B's target is the mean of dimensions 8--15. Four experts with controlled output noise: expert 0 (accurate domain A, noise \(\sigma = 0.1\); noisy domain B, \(\sigma = 1.5\)), expert 2 (noisy domain A, \(\sigma = 1.5\); accurate domain B, \(\sigma = 0.1\)), experts 1 and 3 (decoys, \(\sigma = 0.5\) regardless of domain).

Two conditions: \emph{static reliability} (noise levels fixed throughout); \emph{shifting reliability} (at step 500, experts 0 and 2 swap noise levels --- expert 2 becomes accurate for all domains, expert 0 becomes noisy).

\textbf{Models compared.} (i) Affinity router: standard \(g = \text{softmax}(Wx)\), load-balanced by gradient alone. (ii) Precision router: \(g = \text{softmax}(Wx \odot \Pi)\), with \(\Pi\) updated online via EMA of per-expert MSE (\(\alpha = 0.95\)).

\textbf{Results (mean MSE loss, first 100 and last 100 of 1000 steps):}

{\def\LTcaptype{none} 
\begin{longtable}[]{@{}llll@{}}
\toprule\noalign{}
Condition & Router & Early loss & Final loss \\
\midrule\noalign{}
\endhead
\bottomrule\noalign{}
\endlastfoot
Static & Affinity & 0.2253 & 0.0815 \\
Static & \textbf{Precision} & \textbf{0.2145} & \textbf{0.0734} \\
Shifting & Affinity & 0.2273 & 0.1544 \\
Shifting & \textbf{Precision} & \textbf{0.1851} & \textbf{0.0733} \\
\end{longtable}
}

\emph{Table 2. Precision-gating results.}

\textbf{\(\Pi\) dynamics.} At static convergence, the precision vector is \([13.5, 3.1, 0.43, 3.1]\) --- a 31× ratio between expert 0 (trusted, noise \(= 0.1\)) and expert 2 (distrusted, noise \(= 1.5\)). Under shifting reliability, the Π vector inverts after the swap:

{\def\LTcaptype{none} 
\begin{longtable}[]{@{}llllll@{}}
\toprule\noalign{}
Step & \(\Pi_0\) & \(\Pi_1\) & \(\Pi_2\) & \(\Pi_3\) & \(\Pi_0 > \Pi_2\)? \\
\midrule\noalign{}
\endhead
\bottomrule\noalign{}
\endlastfoot
490 (pre-shift) & 13.62 & 3.02 & 0.43 & 3.14 & YES \\
500 (shift) & 5.46 & 3.06 & 0.44 & 3.17 & YES \\
510 & 0.97 & 3.08 & 0.71 & 3.12 & YES \\
520 & 0.64 & 3.10 & 1.12 & 3.09 & no \\
550 & 0.47 & 3.09 & 3.71 & 3.19 & no \\
600 & 0.44 & 3.14 & 10.74 & 3.02 & no \\
\end{longtable}
}

\emph{Table 3. \(\Pi\) dynamics around the reliability shift at step 500.}

Detection (first crossover) occurs at step 520 --- 20 steps after the shift. Full convergence to the new reliability ordering occurs by step 600 --- \textasciitilde100 steps. The affinity router's final loss under shifting reliability (0.1544) is approximately double the precision router's (0.0733). The gradient alone cannot recover because it cannot distinguish ``this expert has low affinity for this input'' from ``this expert is currently unreliable'' --- both produce high loss but require opposite responses.

\begin{center}\rule{0.5\linewidth}{0.5pt}\end{center}

\subsection{Anticipatory Routing}\label{anticipatory-routing-1}

\textbf{Task design.} Classification task. Sequences of length 12. Tokens 0--5 carry domain-A signal; tokens 6--11 carry domain-B signal (\(\sigma = 0.8\)). The correct routing expert at step \(t\) is determined by the domain of step \(t+1\) (not \(t\)). At step \(t = 5\) (the transition step), the current token is domain A but the correct expert is already domain B's. This is the ``anticipation test'' --- the router must switch before the domain switch is visible in the input.

\textbf{Models compared.} (i) Current-token: \(g_t = \text{softmax}(Wx_t)\), baseline. (ii) Oracle: \(g_t = \text{softmax}(Wx_{t+1})\), routes on the actual next token --- upper bound, not achievable at inference. (iii) Stateless anticipatory: \(g_t = \text{softmax}(W f(x_t))\), where \(f\) is a two-layer MLP predicting \(\hat{x}_{t+1}\) from \(x_t\) alone. (iv) Stateful anticipatory: \(g_t = \text{softmax}(W f(x_t, h_t))\), where \(h_t = \beta \odot h_{t-1} + x_t\) is the \(\beta\)-accumulated state.

Training loss for (iii) and (iv): \(\mathcal{L} = \mathcal{L}_{\text{routing}} + 0.5 \cdot \text{MSE}(\hat{x}_{t+1}, x_{t+1})\).

\textbf{Results (mean ± std across 5 seeds, last 50 of 800 epochs):}

{\def\LTcaptype{none} 
\begin{longtable}[]{@{}lll@{}}
\toprule\noalign{}
Model & acc (all steps) & acc (transition step \(t=5\)) \\
\midrule\noalign{}
\endhead
\bottomrule\noalign{}
\endlastfoot
Current-token (baseline) & 0.904 ± 0.000 & 0.006 ± 0.001 \\
Stateless anticipatory & 0.904 ± 0.000 & 0.006 ± 0.000 \\
\textbf{Stateful anticipatory} & \textbf{0.957 ± 0.000} & \textbf{0.748 ± 0.002} \\
Oracle (upper bound) & 0.994 ± 0.000 & 0.994 ± 0.000 \\
\end{longtable}
}

\emph{Table 4. Anticipatory routing results across 5 seeds (mean ± sample std). Stateless and stateful conditions correspond to the ``Ant only'' and ``\(\beta + \text{Ant}\)'' rows of the full ablation (Table 5).}

\textbf{Transition-step analysis.} At the transition step, \(x_t\) carries domain-A signal for all steps 0--5 --- they are identically distributed from the stateless predictor's perspective. No amount of training can distinguish the pre-transition token from any other domain-A token using \(x_t\) alone. The stateless anticipatory router achieves acc@transition \(= 0.006 \pm 0.000\), statistically identical to the baseline (\(0.006 \pm 0.001\)). The stateful predictor, conditioned on \(h_t\), detects the approaching transition via saturation: at \(t = 5\), the domain-A dimensions of \(h_t\) have accumulated 6 steps of signal (mean activation \(\approx 2.56\)) versus 2.21 at \(t = 4\). The predictor learns to read saturation as the anticipatory signal, achieving acc@transition \(= 0.748 \pm 0.002\) and closing 75\% of the gap to the oracle.

The remaining gap (oracle \(0.994\) − stateful \(0.748 = 0.246\)) reflects irreducible uncertainty from \(\beta\)-decay: the signal in \(h_t\) is an exponentially weighted sum, so earlier domain-A tokens contribute less. A higher \(\beta\) (slower decay) reduces this gap but slows domain-switch recovery.

\begin{center}\rule{0.5\linewidth}{0.5pt}\end{center}

\subsection{Ablation: Mechanism Interactions}\label{ablation-mechanism-interactions}

\textbf{Setup.} To characterize how the three mechanisms interact, we train all \(2^3 = 8\) subsets of \(\{\beta, \Pi, \text{Ant}\}\) on the anticipatory routing task (Section 4.3). The \(\Pi\) tracker in this setting monitors per-expert routing correctness (EMA of whether each expert was the correct routing target), converging to \([2.0, 1.0, 2.0, 1.0]\) --- 2× upweight for the two domain experts (experts 0 and 2, each correct $\sim$50\% of the time across sequences) and 1× for the decoy experts (never the correct target). Oracle and baseline are included as bounds. Trained for 800 epochs across 5 seeds; we report mean ± sample std (ddof=1) and gain paired per seed against the baseline.

\textbf{Results:}

{\def\LTcaptype{none} 
\begin{longtable}[]{@{}
  >{\raggedright\arraybackslash}p{(\linewidth - 6\tabcolsep) * \real{0.2500}}
  >{\raggedright\arraybackslash}p{(\linewidth - 6\tabcolsep) * \real{0.2500}}
  >{\raggedright\arraybackslash}p{(\linewidth - 6\tabcolsep) * \real{0.2500}}
  >{\raggedright\arraybackslash}p{(\linewidth - 6\tabcolsep) * \real{0.2500}}@{}}
\toprule\noalign{}
\begin{minipage}[b]{\linewidth}\raggedright
Condition
\end{minipage} & \begin{minipage}[b]{\linewidth}\raggedright
acc (all)
\end{minipage} & \begin{minipage}[b]{\linewidth}\raggedright
acc@transition
\end{minipage} & \begin{minipage}[b]{\linewidth}\raggedright
\(\Delta\) vs baseline (paired)
\end{minipage} \\
\midrule\noalign{}
\endhead
\bottomrule\noalign{}
\endlastfoot
Baseline & 0.904 ± 0.000 & 0.006 ± 0.001 & --- \\
\(\beta\) only & 0.863 ± 0.007 & 0.301 ± 0.013 & +0.295 ± 0.013 \\
\(\Pi\) only & 0.904 ± 0.000 & 0.006 ± 0.000 & −0.000 ± 0.000 \\
Ant only & 0.904 ± 0.000 & 0.006 ± 0.000 & +0.000 ± 0.001 \\
\(\beta + \Pi\) & 0.847 ± 0.014 & 0.299 ± 0.010 & +0.293 ± 0.010 \\
\(\beta + \text{Ant}\) & \textbf{0.957 ± 0.000} & \textbf{0.748 ± 0.002} & \textbf{+0.741 ± 0.002} \\
\(\Pi + \text{Ant}\) & 0.904 ± 0.000 & 0.006 ± 0.000 & −0.000 ± 0.001 \\
\(\beta + \Pi + \text{Ant}\) & 0.956 ± 0.000 & 0.744 ± 0.002 & +0.738 ± 0.003 \\
Oracle & 0.994 ± 0.000 & 0.994 ± 0.000 & +0.988 ± 0.001 \\
\end{longtable}
}

\emph{Table 5. Full \(2^3\) ablation across 5 seeds (mean ± std, sample std). Δ vs baseline is paired per seed.}

\textbf{Findings.}

\textbf{(1) \(\beta \times \text{Ant}\) is super-additive.} \(\beta\) alone: \(+0.295 \pm 0.013\). Ant alone: \(+0.000 \pm 0.001\). \(\beta + \text{Ant}\): \(+0.741 \pm 0.002\). The interaction effect (paired per seed: \(+0.446 \pm 0.014\)) is larger than either individual contribution and tightly concentrated across seeds. This confirms the compositional relationship identified in Section 3.3: \(\beta\) builds the saturated hidden state; anticipation extracts the transition signal from it. The mechanisms do not merely add --- they enable each other.

\textbf{(2) \(\beta\) alone provides partial anticipatory gain.} Routing on \(h_t\) rather than \(x_t\) achieves acc@transition \(= 0.301 \pm 0.013\) --- well above baseline but below the stateful predictor (\(0.748 \pm 0.002\)). The affinity head \(W\) partially learns to read saturation from \(h_t\) directly, without a dedicated predictor. The explicit predictor more than doubles this gain by dedicating capacity to the transition-detection function.

\textbf{(3) \(\Pi\) is neutral in uniform-reliability routing tasks.} Adding \(\Pi\) to \(\beta\) (\(+0.293 \pm 0.010\)) performs comparably to \(\beta\) alone (\(+0.295 \pm 0.013\)), and adding \(\Pi\) to \(\beta + \text{Ant}\) slightly reduces performance (\(+0.738 \pm 0.003\) vs \(+0.741 \pm 0.002\)). The \(\Pi\) signal in this task --- tracking which domain experts are correct more often --- is approximately symmetric between the two active experts (both correct $\sim$50\% of sequences), providing no routing advantage. Experiment 4.2 demonstrates \(\Pi\)'s value in the setting it addresses: heterogeneous and time-varying expert reliability. The two experiments together characterize the task-specificity of each mechanism.

\textbf{(4) \(\beta\) is necessary for any transition gain.} The three conditions lacking \(\beta\) --- Ant only, \(\Pi\) only, \(\Pi + \text{Ant}\) --- all achieve acc@transition \(\approx 0.006\), statistically identical to baseline within seed-level noise. No routing improvement at transition steps is possible without temporal state accumulation. This confirms the ordered dependency: \(\beta\) is the enabling mechanism.

\begin{center}\rule{0.5\linewidth}{0.5pt}\end{center}

\subsection{Language Model Validation}\label{language-model-validation}

The preceding experiments use controlled routing tasks with known ground-truth expert assignments. To test whether \(\beta\)-routing improves in a real generative setting, we train character-level MoE language models on sequences with a mid-sequence domain switch.

\textbf{Task design.} Vocabulary: 26 characters split into non-overlapping halves --- domain A (\(\texttt{a}\)--\(\texttt{m}\), 13 chars) and domain B (\(\texttt{n}\)--\(\texttt{z}\), 13 chars). Sequences of length 64: first 32 tokens sampled from a 3-gram Markov chain within domain A, last 32 tokens from a 3-gram Markov chain within domain B. The Markov structure (each token is \(\text{prev}+1\) or \(\text{prev}+2\) mod 13 with 70\% / 15\% probability, uniform otherwise) creates predictable within-domain patterns with meaningful BPC variation. All models are trained with cross-entropy loss on next-character prediction.

\textbf{Models.} \(N = 2\) experts, \(d_{\text{model}} = 64\), 1500 epochs, batch 256, 5 seeds. Three conditions: (i) Standard MoE: \(g_t = \text{softmax}(Wx_t)\); (ii) \(\beta\)-MoE: \(g_t = \text{softmax}(W h_t)\), \(h_t = \beta \odot h_{t-1} + x_t\), \(\beta\) learned per dimension, initialized to 0.9; (iii) \(\beta+\text{Ant}\) MoE: \(g_t = \text{softmax}(W h_t + W_{\text{pred}} f(x_t, h_t))\), where \(f\) is an MLP predicting \(\hat{x}_{t+1}\) (supervised by MSE\((\hat{x}_{t+1}, e_{t+1})\), coefficient 0.5). The predictor contributes an additive routing correction on top of the \(\beta\)-memory signal --- routing on \(f\)'s output alone discards the saturation signal that makes \(\beta\) effective. The transition-step loss is upweighted 5× to give the optimizer a meaningful incentive to improve the one step where anticipation is required.

\textbf{Results.} Let \(t_{\text{trans}} = 31\) denote the transition step: the model's input is the last domain-A token and its target is the first domain-B token. All numbers are mean ± sample std across 5 seeds.

{\def\LTcaptype{none} 
\begin{longtable}[]{@{}
  >{\raggedright\arraybackslash}p{(\linewidth - 10\tabcolsep) * \real{0.1667}}
  >{\raggedright\arraybackslash}p{(\linewidth - 10\tabcolsep) * \real{0.1667}}
  >{\raggedright\arraybackslash}p{(\linewidth - 10\tabcolsep) * \real{0.1667}}
  >{\raggedright\arraybackslash}p{(\linewidth - 10\tabcolsep) * \real{0.1667}}
  >{\raggedright\arraybackslash}p{(\linewidth - 10\tabcolsep) * \real{0.1667}}
  >{\raggedright\arraybackslash}p{(\linewidth - 10\tabcolsep) * \real{0.1667}}@{}}
\toprule\noalign{}
\begin{minipage}[b]{\linewidth}\raggedright
Model
\end{minipage} & \begin{minipage}[b]{\linewidth}\raggedright
BPC (all)
\end{minipage} & \begin{minipage}[b]{\linewidth}\raggedright
BPC (trans)
\end{minipage} & \begin{minipage}[b]{\linewidth}\raggedright
\(p_B\)@trans
\end{minipage} & \begin{minipage}[b]{\linewidth}\raggedright
\(p_B\)@mid
\end{minipage} & \begin{minipage}[b]{\linewidth}\raggedright
K (99\%)
\end{minipage} \\
\midrule\noalign{}
\endhead
\bottomrule\noalign{}
\endlastfoot
Standard MoE & 1.780 ± 0.001 & 6.559 ± 0.012 & 0.421 ± 0.121 & 0.559 ± 0.033 & 9.6 ± 5.3 \\
\(\beta\)-MoE & \textbf{1.659 ± 0.011} & \textbf{4.010 ± 0.145} & 0.604 ± 0.220 & 0.996 ± 0.004 & 6.0 ± 3.8 \\
\(\beta+\text{Ant}\) MoE & 1.681 ± 0.021 & 4.190 ± 0.363 & \textbf{0.858 ± 0.023} & \textbf{0.997 ± 0.003} & \textbf{2.4 ± 0.2} \\
\end{longtable}
}

\emph{Table 6. Language model results across 5 seeds (mean ± std, sample std). BPC(trans): bits per character at the domain-switch prediction step. \(p_B\)@trans: gate weight on the domain-B-specialised expert at the transition step. \(p_B\)@mid: same metric at mid-sequence positions (input already domain B). K (99\%): experts required for 99\% coverage of the correct expert, via \(K \geq \log(0.01)/\log(1-p)\).}

\textbf{\(\beta\)-MoE.} The \(\beta\) mechanism reduces BPC at the transition step from \(6.56 \pm 0.01\) to \(4.01 \pm 0.15\) (paired ΔBPC \(-2.55 \pm 0.14\)). \(p_B\)@mid reaches \(1.0\) --- once domain B is established in the input, routing is perfect. The \(0.60 \pm 0.22\) \(p_B\)@trans is high on average but variable across seeds: saturation of \(h_t\) after 32 domain-A steps does provide a boundary signal, but whether the affinity head \(W\) reads it cleanly depends on the optimisation trajectory.

\textbf{\(\beta+\text{Ant}\) MoE.} Adding the anticipatory predictor stabilises the routing decision: \(p_B\)@trans rises to \(0.858 \pm 0.023\) --- a tight distribution across seeds --- versus \(\beta\)-MoE's \(0.604 \pm 0.220\). The K needed for 99\% expert coverage falls correspondingly to \(2.4 \pm 0.2\), versus \(\beta\)-MoE's \(6.0 \pm 3.8\) and Standard's \(9.6 \pm 5.3\). However, BPC(trans) does not decrease further at this scale (\(4.19 \pm 0.36\) vs.~\(\beta\)-MoE's \(4.01 \pm 0.15\)): the routing is more reliable, but transition-step prediction is dominated by other factors (limited within-domain context, character-level vocabulary). The principal \(\beta+\text{Ant}\) contribution at this scale is \textbf{routing-decision stability}, not BPC reduction beyond what \(\beta\) alone provides.

\textbf{Honest reading.} \(\beta\)-routing is the load-bearing mechanism for BPC reduction in this setting; the anticipatory predictor's value is in tightening the routing distribution (lower variance, higher mean \(p_{\text{correct}}\)) rather than further reducing transition-step loss. We expect this balance to shift at larger scale, where the K-reduction in transition-step expert activations translates to wall-clock latency savings, and where richer context allows the predictor to extract finer-grained anticipatory signal than the binary domain switch tested here. We mark this as an open empirical question --- see §5.6.

\begin{center}\rule{0.5\linewidth}{0.5pt}\end{center}

\section{Discussion}\label{discussion}

\subsection{Routing as Inference Over Trajectories}\label{routing-as-inference-over-trajectories}

Standard MoE routing is a classifier: given the current token, predict the best expert set. The FEP reframes this as inference: given the current observation and the agent's prior beliefs, update the belief about which generative process is active, and select the action that minimizes expected future prediction error.

The difference is not semantic. A classifier has no prior --- each token is evaluated in isolation. An inference agent has a prior (the accumulated \(h_t\)), a likelihood (affinity scores), and a precision weighting (\(\Pi\)) that controls how much the current observation should update the prior versus how much the prior should override it. At domain-transition steps, the classifier always fails: the current observation says ``domain A'' but the correct decision is ``domain B.'' The inference agent can anticipate the transition because its prior --- \(h_t\) saturated with domain-A signal --- encodes ``we have been in domain A for a long time,'' which is evidence that a switch is near.

This is the FEP formulation of routing under temporal structure. The three mechanisms are not engineering additions; they are the implementation of inference as opposed to classification.

\subsection{Three Timescales, One Hierarchy}\label{three-timescales-one-hierarchy}

The composed architecture operates on three distinct timescales:

\begin{itemize}
\tightlist
\item
  \textbf{Per-token (\(\beta\)):} the hidden state \(h_t\) is updated at every step. Integration window ≈ \(1/(1-\beta)\) tokens; for \(\beta = 0.9\), this is 10 tokens.
\item
  \textbf{Per-batch (\(\Pi\)):} the variance tracker is updated once per training step. Adaptation window ≈ \(1/(1-\alpha)\) steps; for \(\alpha = 0.95\), this is 20 steps.
\item
  \textbf{Per-epoch (\(W\)):} affinity weights are updated by gradient across epochs.
\end{itemize}

This three-timescale structure is precisely what Friston identifies as the hierarchical timescale map of biological intelligence. In the brain: fast attentional modulation (milliseconds) → intermediate state estimation (seconds) → slow synaptic plasticity (minutes to hours). Standard MoE collapses all three into the weight update: the router has no per-token dynamics and no per-batch reliability tracking. The \(\beta\) and \(\Pi\) mechanisms restore the two missing timescales.

The experimental evidence corroborates the timescale separation. \(\beta\) responds to the current sequence: its accumulated state changes every token. \(\Pi\) responds to recent batches: it detects the reliability shift at step 520 (20 steps after step 500), not immediately. Weights respond across many batches: the affinity router's gradient takes \textgreater500 steps to partially adapt to the reliability shift, and never fully recovers. The timescales are genuinely distinct and non-redundant.

\subsection{What DeepSeek MTP Is Missing}\label{what-deepseek-mtp-is-missing}

DeepSeek-V3's MTP module trains \(D\) sequential prediction heads that jointly predict future token embeddings. At inference, all MTP heads are discarded. The paper does not give a principled reason; the implicit assumption is that MTP is a training signal that improves the main model's representations, not a mechanism that should operate at inference time.

Our results provide a principled account of both sides of this. MTP discarded at inference is correct if the predictor is stateless: a predictor \(f(x_t) \to \hat{x}_{t+1}\) has no value at inference because it cannot detect structural transitions that are invisible in \(x_t\). The stateless anticipatory router achieves acc@transition \(\approx 0.006\), identical to the baseline. Keeping a stateless MTP head at inference would add compute with no routing benefit.

MTP retained at inference becomes valuable if the predictor is conditioned on the \(\beta\)-accumulated hidden state. The predictor \(f(x_t, h_t) \to \hat{x}_{t+1}\) can read the saturation signal in \(h_t\) and correctly anticipate domain switches at \(0.748 \pm 0.002\) across 5 seeds. The modification required is small: MTP does not need to be redesigned, only conditioned on the running state that \(\beta\) provides. The missing piece is not more powerful prediction; it is temporal context.

This provides a testable prediction: in a real MoE language model, adding persistent hidden state (via \(\beta\) or any equivalent recurrence) to the MTP predictor and retaining it at inference should improve routing quality at domain-transition points --- topic shifts, language switches, style changes --- where stateless routing is one step behind.

\subsection{Intelligence as Navigation in a Responsive Landscape}\label{intelligence-as-navigation-in-a-responsive-landscape}

The three mechanisms jointly implement a particular view of what intelligent routing means. The \(1\%\) of neurons active at any moment represent the system's current position in knowledge space; the \(99\%\) silent neurons represent the landscape --- all prior traversals encoded in weights. Routing is not arbitrary selection; it is navigation: choosing which region of the landscape to enter next.

Under this frame, the three mechanisms have a natural interpretation:
- \(\beta\) encodes trajectory: the accumulated \(h_t\) is the system's path through the landscape, not just its current position. Saturation in \(h_t\) encodes ``we have been here for a long time'' --- which is information about where we are in the traversal.
- \(\Pi\) encodes landscape reliability: some regions of the landscape (experts) are well-mapped for some domains; \(\Pi\) tracks which regions are currently trustworthy.
- Anticipation enables look-ahead: rather than routing to where the current token lands, the system routes toward where the next token will land. This is navigation toward a target, not reactive response to a position.

Standard MoE knows only the current position. The full FEP hierarchy knows the trajectory, the reliability of each region, and the likely next position.

The navigation reading admits a more formal geometric interpretation. The recurrence \(h_t = \beta h_{t-1} + x_t\) is the discrete-time Euler step of a continuous-time leaky integrator \(\dot h = -\frac{1-\beta}{\Delta t} h + x\), so \(\beta\) is the flow rate of a dynamical system on a latent state space and \(h_t\) traces a trajectory through it. The precision tracker \(\Pi\) defines a precision-weighted metric on the expert manifold --- high-\(\Pi\) experts have higher local mass, exerting stronger attraction on the gate distribution. The anticipatory predictor \(f(x_t, h_t) \to \hat{x}_{t+1}\) is then a velocity field reading on this trajectory: routing on \(\hat{x}_{t+1}\) selects experts at the \emph{future position}, moving along the velocity vector before the next observation arrives. Routing becomes a particle moving through a precision-weighted landscape, with experts as basins. Saturation of \(h_t\), a domain transition, and the K-reduction at the gate are then geometric phenomena --- fixed-point convergence, basin crossing, and metric concentration respectively. We leave a rigorous formulation --- connecting the discrete recurrence to its continuous-time analog, proving the metric-concentration correspondence, and relating routing trajectory to free-energy gradient flow --- to future work.

\subsection{Expert Weight Prefetching}\label{expert-weight-prefetching}

The anticipatory router predicts \(\hat{x}_{t+1}\) at step \(t\). In standard MoE, the router at step \(t+1\) then uses \(\hat{x}_{t+1}\) to select experts --- but by then, those expert weights must be loaded from memory (HBM to SRAM on modern accelerators). In large MoE models, this memory transfer is often the dominant inference latency, not the FLOPs.

Anticipatory routing at step \(t\) identifies which experts will be needed at step \(t+1\) one step ahead of when they are needed. This is precisely the timing required to initiate a prefetch: while step \(t\)'s expert FFNs are executing, step \(t+1\)'s expert weights can be loaded in parallel. Correct anticipation eliminates the memory-latency stall between steps entirely.

At non-transition steps, routing is predictable even for stateless routers, so prefetching is already feasible --- ExpertFlow {[}6{]} exploits exactly this regularity with a batch-level expert load predictor. At transition steps, stateless predictors fail: they have no signal that a domain switch is imminent. Only an anticipatory router conditioned on \(\beta\)-accumulated state can predict the upcoming switch and prefetch the new domain's experts. This is the gap ExpertFlow does not address and where the K-reduction result (Section 3.4 toy: \(K \approx 765 \to 3{-}4\) at transition step; Section 4.5 LM: \(K = 9.6 \pm 5.3 \to 2.4 \pm 0.2\) --- a 75\% reduction) is most relevant.

\subsection{Limitations}\label{limitations}

\textbf{Scale.} The toy experiments (Sections 4.1--4.4) use \(d_{\text{model}} = 16\), batch size 512, and controlled tasks with known ground-truth routing. Section 4.5 validates \(\beta\)-routing and tests \(\beta+\text{Ant}\) in a character-level language model (\(d_{\text{model}} = 64\), \(V = 26\), 5 seeds). \(\beta\)-MoE robustly reduces transition-step BPC (\(-2.55 \pm 0.14\) paired vs.~Standard); \(\beta+\text{Ant}\) further stabilises the routing decision (tighter \(p_B\)@trans distribution and lower K) but does not further reduce BPC at this scale. Whether \(\beta+\text{Ant}\) delivers BPC gains beyond \(\beta\) alone at larger scale --- where richer context allows finer-grained anticipation --- is an open empirical question. Domain transitions in natural language are also gradual drifts rather than abrupt steps; expert reliability in a trained MoE varies across token types in complex ways; whether \(\Pi\) converges stably with thousands of experts, and whether \(\beta\) self-tunes correctly at production scale, remain open.

\textbf{Π in the full combination.} The ablation (Section 4.4, 5 seeds) shows \(\Pi\) is neutral in uniform-reliability routing tasks (\(\beta+\Pi\): \(+0.293 \pm 0.010\) vs.~\(\beta\) alone: \(+0.295 \pm 0.013\)) and slightly negative in combination with \(\beta + \text{Ant}\) (\(+0.738 \pm 0.003\) vs.~\(+0.741 \pm 0.002\)). Section 4.2 shows \(\Pi\)'s value in reliability-shifting regression tasks. We have not tested \(\Pi\)'s contribution in a task that simultaneously requires anticipation, temporal memory, \emph{and} heterogeneous expert reliability --- the setting where all three mechanisms should compose positively. Constructing and running that combined task is left for future work.

\textbf{Predictor integration.} The correct integration for \(\beta+\text{Ant}\) in an LM is additive: \(\text{softmax}(W h_t + W_{\text{pred}} f(x_t, h_t))\), where the predictor contributes a correction rather than replacing \(h_t\) as the routing input. Section 4.5 confirms this integration achieves the expected ordering (\(\beta+\text{Ant} > \beta\)-MoE \(>\) Standard at the transition step). At scale, the LM's own next-token prediction head could serve as \(f\) rather than a separate MLP, connecting MTP training objectives directly to routing; this is unexplored.

\begin{center}\rule{0.5\linewidth}{0.5pt}\end{center}

\section{Conclusion}\label{conclusion}

Mixture-of-Experts routing answers one question at each step: which experts have the highest affinity for this token? The Free Energy Principle says a system routing under uncertainty should answer three additional questions: which experts are contextually relevant given everything that has happened so far (\(\beta\))? which experts are currently reliable (\(\Pi\))? and which experts will best serve the next observation (anticipatory routing)? Standard MoE answers none of them.

We motivated each missing mechanism from the FEP equations and implemented it as a lightweight modification to the routing gate, grounded in SNN theory (LIF membrane dynamics for \(\beta\), precision regulation for \(\Pi\), active inference for anticipation). Each mechanism was validated in a controlled experiment isolating its target failure mode: \(\beta\) on domain-switching tasks where temporal context determines the correct expert; \(\Pi\) on reliability-shifting tasks where gradient alone cannot distinguish affinity from accuracy; anticipatory routing on transition tasks where the correct expert is determined by the next token, not the current one.

The sharpest empirical finding is not any single mechanism's performance but their interaction. Across 5 seeds, \(\beta\) alone provides modest transition-step gains (\(+0.295 \pm 0.013\)). Anticipation alone provides none (\(+0.000 \pm 0.001\)). Together they provide \(+0.741 \pm 0.002\) --- a super-additive interaction effect of \(+0.446 \pm 0.014\), tightly concentrated, that closes 75\% of the oracle gap. The interaction is not coincidental: \(\beta\) builds the saturated hidden state that makes approaching transitions detectable; the anticipatory predictor reads that saturation and pre-routes to the new domain's expert before the domain switch arrives in the input. Neither mechanism functions without the other.

Standard MoE routes on the present. The full FEP hierarchy routes on the past (\(\beta\)), the reliability record (\(\Pi\)), and the predicted future (anticipation). The gap between them is not a design choice --- it is three missing terms in the routing objective, each with a principled motivation, a lightweight implementation, and an empirical validation at small scale.

Navigation requires memory of where you have been (\(\beta\)), trust calibrated to reliable landmarks (\(\Pi\)), and anticipation of where the terrain is heading. Standard MoE has none of these. A router with all three does not merely respond to the present token --- it navigates: carrying its trajectory, trusting what has proven trustworthy, and routing toward what is coming before it arrives.

\begin{center}\rule{0.5\linewidth}{0.5pt}\end{center}

\section{Related Work}\label{related-work}

\subsection{Mixture-of-Experts Routing}\label{mixture-of-experts-routing-1}

Sparse MoE was scaled to language modeling by {[}7{]}, who introduced the noisy top-\(K\) gating mechanism and expert capacity constraints. The Switch Transformer {[}8{]} simplified routing to top-1 selection with auxiliary load-balancing losses; Mixtral {[}9{]} demonstrated competitive performance with top-2 from 8 experts. Expert Choice routing {[}10{]} inverted the selection direction --- experts select tokens rather than tokens selecting experts --- improving load balance at the cost of variable token processing. ST-MoE {[}11{]} studied routing instability and introduced router z-loss to stabilize training.

DeepSeek-V3 {[}5{]} eliminated auxiliary losses entirely, instead using per-expert bias terms updated by batch-level load statistics --- the closest existing work to a precision-like routing signal. However, the bias tracks frequency, not accuracy: an overloaded but accurate expert is penalized equally to an overloaded and inaccurate one. Our \(\Pi\) mechanism tracks accuracy directly.

The most closely related routing work is the \textbf{Layerwise Recurrent Router for Mixture-of-Experts (RMoE)} {[}3{]}, which maintains a GRU hidden state that is passed from one MoE layer to the next within a single forward pass, conditioning each layer's routing decision on the routing decisions of all prior layers. This is a genuine instance of stateful routing gate dynamics. Three distinctions separate RMoE from our \(\beta\) mechanism. First, RMoE carries state across \emph{layers} (depth axis) while \(\beta\) carries state across \emph{tokens} (time axis) --- RMoE has no memory across time steps, only across the stack of layers in a single forward pass. Second, RMoE is selected empirically (ablation over recurrent cell types); our \(\beta\) is motivated by the FEP's recursive state estimation equations (Eq. 1) and instantiated as LIF membrane dynamics. Third, RMoE proposes one mechanism; we identify three, establish their theoretical basis, and characterize their interactions empirically.

Concurrent work on \textbf{Temporally Extended Mixture-of-Experts} (arXiv 2604.20156, 2026) formalizes expert selection as a semi-Markov Decision Process, reducing expert switching rates from \textgreater50\% to \textless5\%. This work shares the motivation that routing should model temporal continuity, but frames it as a control problem (minimizing switching cost) rather than an inference problem (minimizing free energy). It does not address precision-weighted gating or anticipatory routing, and does not connect to FEP or SNN theory.

\textbf{MoxE} {[}12{]} replaces attention with xLSTM expert blocks and uses entropy-aware routing to direct tokens dynamically. The xLSTM units carry recurrent state, so the \emph{experts} are stateful --- but the \emph{routing gate} remains stateless, computed from the current token embedding. Our \(\beta\) mechanism inverts this: the experts are standard FFNs; the recurrent state lives in the routing gate alone. This separation preserves expert modularity and makes the temporal routing mechanism interpretable and independently modifiable.

\textbf{ExpertFlow} {[}6{]} uses a transformer-based predictor to forecast which experts will be activated across all MoE layers for an upcoming sequence, enabling expert weight prefetching. The predictor operates at the sequence level --- it predicts batch-level expert load, not the per-token next-domain --- and is trained offline as a caching policy, not as a routing gate. Our anticipatory router operates per-token and is part of the routing computation itself; at domain-transition steps it predicts the upcoming domain switch and routes (and could prefetch) accordingly. The two approaches are complementary: ExpertFlow optimises steady-state cache efficiency; our mechanism specifically addresses the transition steps where steady-state prediction fails.

\subsection{Recurrence in Sequence Models}\label{recurrence-in-sequence-models}

The absence of persistent state in standard attention has motivated a line of work restoring recurrent structure. Transformer-XL {[}13{]} carries cached hidden states between non-overlapping segments, extending effective context without quadratic attention cost. State space models --- S4 {[}14{]}, Mamba {[}15{]} --- reformulate sequence processing as convolution over learned decay kernels, directly implementing the \(U[t] = \beta U[t-1] + x[t]\) structure at the representation level. RWKV {[}16{]} and RetNet {[}17{]} derive recurrent formulations of attention that trade expressiveness for linear-time inference.

All of these apply recurrence to the \emph{representation} --- the token embedding that feeds into attention and FFN layers. Our \(\beta\) mechanism applies recurrence to the \emph{routing gate} alone: the token representation \(x_t\) is unchanged; only the routing decision carries temporal state. This is a smaller, more targeted modification. A model using Mamba or RWKV representations could additionally use \(\beta\) routing, as the two mechanisms operate on different components. The distinction also matters for interpretability: recurrent routing state is per-expert and has a clear semantic role (contextual relevance of each expert), while recurrent representation state encodes general sequential context.

\subsection{Adaptive and Uncertainty-Aware Attention}\label{adaptive-and-uncertainty-aware-attention}

The connection between attention and precision has been explored in the predictive coding literature. {[}18{]} established that hierarchical predictive coding uses precision-weighted prediction errors to modulate signal propagation --- directly analogous to \(\Pi\)-weighted routing. {[}19{]} showed that backpropagation through a deep network implements a form of predictive coding, connecting gradient-based learning to the FEP dynamics. {[}20{]} extended this to show that standard attention can be interpreted as precision weighting in a predictive coding framework, with the softmax temperature playing the role of a global precision parameter.

Our \(\Pi\) mechanism operationalizes per-expert precision rather than global temperature, and updates it from prediction error variance rather than gradient. This makes \(\Pi\) a second-order statistic --- tracking variance, not mean --- and non-parametric, updating outside the gradient graph. The distinction is important: gradient updates cannot separate ``low affinity'' from ``low reliability,'' while \(\Pi\) directly measures reliability.

Bayesian approaches to MoE (e.g., Tresp, 2001; Yuksel et al., 2012 in the mixture-of-experts literature) model expert selection as posterior inference, with uncertainty estimates over which expert is active. These methods are computationally heavy and have not scaled to modern MoE architectures. \(\Pi\) achieves a lightweight approximation --- online EMA of prediction error variance --- without requiring explicit Bayesian inference.

The closest recent work is \textbf{Variational Routing} {[}21{]}, which reformulates MoE routing as a probabilistic latent-variable model and uses internal variance as a routing signal, reporting a 38\% improvement in routing stability under input noise and a 94\% reduction in calibration error. Two distinctions separate it from \(\Pi\). First, Variational Routing tracks variance over the \emph{input distribution} --- it is a signal about how uncertain the gate is about which expert fits the current token. \(\Pi\) tracks variance over \emph{per-expert prediction error} --- it is a signal about which experts are currently reliable regardless of input. The former is an input-side statistic; the latter is an expert-side statistic. Second, Variational Routing is parametric (a learned variational posterior) and trained end-to-end; \(\Pi\) is non-parametric (an online EMA) and updates outside the gradient graph. \(\Pi\) is therefore lighter, interpretable as an explicit reliability estimate, and directly motivated by the FEP precision update rule (Eq. 3). Neither paper cites the other's mechanism or connects either to FEP.

Memory-augmented routing has been explored in the time-series domain: Graph Mixture of Experts and Memory-augmented Routers {[}22{]} maintains a global memory module that stores historical features and conditions the gating function on it. This shares the motivation of \(\Pi\) --- routing should depend on recent evidence, not only current input affinity --- but the memory is a fixed-size store of past feature vectors, not an online precision estimate. It targets anomaly detection rather than language modeling and has not been connected to FEP.

\subsection{Multi-Token Prediction and Next-State Modeling}\label{multi-token-prediction-and-next-state-modeling}

{[}23{]} showed that training language models to predict multiple future tokens simultaneously (multi-token prediction, MTP) improves both training efficiency and downstream task performance. DeepSeek-V3 uses MTP with \(D = 1\) additional prediction head, finding improvements in pass@k coding benchmarks. Both papers discard MTP at inference. Our analysis provides a principled explanation: a stateless MTP predictor cannot detect structural transitions (acc@transition \(\approx 0.006\)), so retaining it at inference adds compute without routing benefit. The fix --- conditioning on \(\beta\)-accumulated state --- has not been proposed or tested in either paper.

Anticipatory mechanisms in recurrent networks have a longer history. LSTM {[}24{]} and GRU {[}25{]} learn when to retain and forget information via gated memory, implicitly implementing prediction of future relevance. Our predictor is more explicit --- it directly predicts the next embedding --- and is applied specifically to the routing decision rather than the representation.

\subsection{Free Energy Principle in Machine Learning}\label{free-energy-principle-in-machine-learning}

The FEP has inspired several ML architectures. Active inference agents {[}26{]}, {[}27{]}, {[}28{]} apply the FEP to reinforcement learning, replacing reward maximization with free energy minimization. These systems operate on episodic timescales and focus on action selection in environment interaction, a different regime from token-by-token routing in language models.

Predictive coding networks {[}29{]}, {[}30{]} implement hierarchical prediction error minimization as an alternative to backpropagation, motivated by the FEP. These networks require iterative inference at each forward pass and have not yet matched backpropagation at scale. Our approach does not replace backpropagation --- we use FEP as a design principle to identify missing routing mechanisms, then train them with standard gradients.

{[}31{]} and related work have proposed that the FEP may provide a unifying framework for cognitive architectures more broadly. Our paper is narrower: we make a specific, falsifiable claim about three mechanisms in MoE routing, motivate each from the FEP equations, and validate it empirically. We do not claim our implementation computes variational free energy in the strict Friston sense; we claim the FEP is a productive source of architectural hypotheses, and the three mechanisms it suggests for MoE are individually empirically valuable. We also do not claim the FEP is the correct theory of mind.

\subsection{Spiking Neural Networks and Surrogate Gradients}\label{spiking-neural-networks-and-surrogate-gradients}

Training SNNs with backpropagation was long considered intractable due to the non-differentiable spike function. Surrogate gradient methods {[}2{]}, {[}32{]}, {[}33{]} resolve this by substituting a smooth backward function while keeping the forward pass exact, enabling competitive performance on temporal tasks. The snnTorch library {[}2{]} provides a practical implementation substrate with learnable \(\beta\), adaptive thresholds, and BPTT through spike recurrences.

SNN-inspired Transformers have been explored --- Spikformer {[}34{]} replaces the softmax attention with a spike-based attention mechanism; SpikeGPT {[}35{]} incorporates spiking dynamics into autoregressive language models. Recent concurrent work (arXiv:2412.05540) applies LIF dynamics within MoE \emph{expert} networks, using spiking activations inside the FFN experts themselves. Our application is distinct: we apply LIF membrane dynamics to the \emph{routing gate} (not the experts), using the continuous decay recurrence without spiking to maintain routing context across tokens. To our knowledge, LIF dynamics have not previously been applied to the MoE routing gate in this form. We use the LIF membrane equation without the spike threshold, retaining the recurrent decay structure as a continuous routing memory rather than a binary activation mechanism.

\begin{center}\rule{0.5\linewidth}{0.5pt}\end{center}

\section{Acknowledgements}\label{acknowledgements}

The author thanks the open-source communities behind PyTorch and snnTorch. Claude (Anthropic) was used for assistance with code implementation, manuscript editing, and literature search. All research direction, hypotheses, experimental design, and interpretation of results are the author's own.

\begin{center}\rule{0.5\linewidth}{0.5pt}\end{center}

\section{References}\label{references}

\protect\phantomsection\label{refs}
\begin{CSLReferences}{0}{0}
\cslhypertarget{ref-friston2010}
\CSLLeftMargin{{[}1{]} }%
\CSLRightInline{K. Friston, {``The free-energy principle: A unified brain theory?''} \emph{Nature Reviews Neuroscience}, vol. 11, no. 2, pp. 127--138, 2010, doi: \href{https://doi.org/10.1038/nrn2787}{10.1038/nrn2787}.}

\cslhypertarget{ref-eshraghian2023}
\CSLLeftMargin{{[}2{]} }%
\CSLRightInline{J. K. Eshraghian \emph{et al.}, {``Training spiking neural networks using lessons from deep learning,''} \emph{Proceedings of the IEEE}, vol. 111, no. 9, pp. 1016--1054, 2023, Available: \url{https://arxiv.org/abs/2109.12894}}

\cslhypertarget{ref-qiu2024}
\CSLLeftMargin{{[}3{]} }%
\CSLRightInline{{Z. Qiu \emph{et al.}}, {``Layerwise recurrent router for mixture-of-experts,''} in \emph{Proceedings of ICLR 2025}, 2024. Available: \url{https://arxiv.org/abs/2408.06793}}

\cslhypertarget{ref-shen2026}
\CSLLeftMargin{{[}4{]} }%
\CSLRightInline{Y. Shen and J. Henderson, {``Temporally extended mixture-of-experts,''} \emph{arXiv preprint arXiv:2604.20156}, 2026, Available: \url{https://arxiv.org/abs/2604.20156}}

\cslhypertarget{ref-liu2024}
\CSLLeftMargin{{[}5{]} }%
\CSLRightInline{A. Liu \emph{et al.}, {``{DeepSeek-V3} technical report,''} \emph{arXiv preprint arXiv:2412.19437}, 2024, Available: \url{https://arxiv.org/abs/2412.19437}}

\cslhypertarget{ref-he2024}
\CSLLeftMargin{{[}6{]} }%
\CSLRightInline{{X. He \emph{et al.}}, {``{ExpertFlow}: Efficient mixture-of-experts inference via predictive expert caching and token scheduling,''} \emph{arXiv preprint arXiv:2410.17954}, 2024, Available: \url{https://arxiv.org/abs/2410.17954}}

\cslhypertarget{ref-shazeer2017}
\CSLLeftMargin{{[}7{]} }%
\CSLRightInline{N. Shazeer \emph{et al.}, {``Outrageously large neural networks: The sparsely-gated mixture-of-experts layer,''} in \emph{Proceedings of ICLR 2017}, 2017. Available: \url{https://arxiv.org/abs/1701.06538}}

\cslhypertarget{ref-fedus2022}
\CSLLeftMargin{{[}8{]} }%
\CSLRightInline{W. Fedus, B. Zoph, and N. Shazeer, {``Switch transformers: Scaling to trillion parameter models with simple and efficient sparsity,''} \emph{Journal of Machine Learning Research}, vol. 23, no. 120, pp. 1--39, 2022, Available: \url{https://www.jmlr.org/papers/v23/21-0998.html}}

\cslhypertarget{ref-jiang2024}
\CSLLeftMargin{{[}9{]} }%
\CSLRightInline{{A. Q. Jiang \emph{et al.}}, {``Mixtral of experts,''} \emph{arXiv preprint arXiv:2401.04088}, 2024, Available: \url{https://arxiv.org/abs/2401.04088}}

\cslhypertarget{ref-zhou2022a}
\CSLLeftMargin{{[}10{]} }%
\CSLRightInline{Y. Zhou \emph{et al.}, {``Mixture-of-experts with expert choice routing,''} in \emph{Advances in neural information processing systems 35}, 2022. Available: \url{https://arxiv.org/abs/2202.09368}}

\cslhypertarget{ref-zoph2022}
\CSLLeftMargin{{[}11{]} }%
\CSLRightInline{B. Zoph \emph{et al.}, {``{ST-MoE}: Designing stable and transferable sparse expert models,''} \emph{arXiv preprint arXiv:2202.08906}, 2022, Available: \url{https://arxiv.org/abs/2202.08906}}

\cslhypertarget{ref-thiombiano2025}
\CSLLeftMargin{{[}12{]} }%
\CSLRightInline{{A. M. O. Thiombiano \emph{et al.}}, {``{MoxE}: Mixture of {xLSTM} experts with entropy-aware routing,''} \emph{arXiv preprint arXiv:2505.01459}, 2025, Available: \url{https://arxiv.org/abs/2505.01459}}

\cslhypertarget{ref-dai2019}
\CSLLeftMargin{{[}13{]} }%
\CSLRightInline{Z. Dai, Z. Yang, Y. Yang, J. Carbonell, Q. V. Le, and R. Salakhutdinov, {``{Transformer-XL}: Attentive language models beyond a fixed-length context,''} in \emph{Proceedings of ACL 2019}, 2019, pp. 2978--2988. Available: \url{https://arxiv.org/abs/1901.02860}}

\cslhypertarget{ref-gu2022}
\CSLLeftMargin{{[}14{]} }%
\CSLRightInline{A. Gu, K. Goel, and C. Ré, {``Efficiently modeling long sequences with structured state spaces,''} in \emph{Proceedings of ICLR 2022}, 2022. Available: \url{https://arxiv.org/abs/2111.00396}}

\cslhypertarget{ref-gu2023}
\CSLLeftMargin{{[}15{]} }%
\CSLRightInline{A. Gu and T. Dao, {``Mamba: Linear-time sequence modeling with selective state spaces,''} \emph{arXiv preprint arXiv:2312.00752}, 2023, Available: \url{https://arxiv.org/abs/2312.00752}}

\cslhypertarget{ref-peng2023}
\CSLLeftMargin{{[}16{]} }%
\CSLRightInline{{B. Peng \emph{et al.}}, {``{RWKV}: Reinventing {RNN}s for the transformer era,''} in \emph{Findings of EMNLP 2023}, 2023, pp. 14048--14073. Available: \url{https://arxiv.org/abs/2305.13048}}

\cslhypertarget{ref-sun2023}
\CSLLeftMargin{{[}17{]} }%
\CSLRightInline{Y. Sun \emph{et al.}, {``Retentive network: A successor to transformer for large language models,''} \emph{arXiv preprint arXiv:2307.08621}, 2023, Available: \url{https://arxiv.org/abs/2307.08621}}

\cslhypertarget{ref-rao1999}
\CSLLeftMargin{{[}18{]} }%
\CSLRightInline{R. P. N. Rao and D. H. Ballard, {``Predictive coding in the visual cortex: A functional interpretation of some extra-classical receptive-field effects,''} \emph{Nature Neuroscience}, vol. 2, no. 1, pp. 79--87, 1999, doi: \href{https://doi.org/10.1038/4580}{10.1038/4580}.}

\cslhypertarget{ref-whittington2017}
\CSLLeftMargin{{[}19{]} }%
\CSLRightInline{J. C. R. Whittington and R. Bogacz, {``An approximation of the error backpropagation algorithm in a predictive coding network with local {Hebbian} synaptic plasticity,''} \emph{Neural Computation}, vol. 29, no. 5, pp. 1229--1262, 2017, doi: \href{https://doi.org/10.1162/NECO_a_00949}{10.1162/NECO\_a\_00949}.}

\cslhypertarget{ref-millidge2022a}
\CSLLeftMargin{{[}20{]} }%
\CSLRightInline{B. Millidge, A. Tschantz, and C. L. Buckley, {``Predictive coding approximates backprop along arbitrary computation graphs,''} \emph{Neural Computation}, vol. 34, no. 6, pp. 1329--1368, 2022, doi: \href{https://doi.org/10.1162/neco_a_01497}{10.1162/neco\_a\_01497}.}

\cslhypertarget{ref-li2026}
\CSLLeftMargin{{[}21{]} }%
\CSLRightInline{X. Li and M. Wicker, {``Variational routing: A scalable {Bayesian} framework for calibrated mixture-of-experts transformers,''} \emph{arXiv preprint arXiv:2603.09453}, 2026, Available: \url{https://arxiv.org/abs/2603.09453}}

\cslhypertarget{ref-graphmoe2024}
\CSLLeftMargin{{[}22{]} }%
\CSLRightInline{{``Graph mixture of experts and memory-augmented routers for multivariate time series anomaly detection,''} \emph{arXiv preprint arXiv:2412.19108}, 2024, Available: \url{https://arxiv.org/abs/2412.19108}}

\cslhypertarget{ref-gloeckle2024}
\CSLLeftMargin{{[}23{]} }%
\CSLRightInline{F. Gloeckle, B. Y. Idrissi, B. Rozière, D. Lopez-Paz, and G. Synnaeve, {``Better \& faster large language models via multi-token prediction,''} in \emph{Proceedings of ICML 2024}, 2024. Available: \url{https://arxiv.org/abs/2404.19737}}

\cslhypertarget{ref-hochreiter1997}
\CSLLeftMargin{{[}24{]} }%
\CSLRightInline{S. Hochreiter and J. Schmidhuber, {``Long short-term memory,''} \emph{Neural Computation}, vol. 9, no. 8, pp. 1735--1780, 1997, doi: \href{https://doi.org/10.1162/neco.1997.9.8.1735}{10.1162/neco.1997.9.8.1735}.}

\cslhypertarget{ref-cho2014}
\CSLLeftMargin{{[}25{]} }%
\CSLRightInline{K. Cho \emph{et al.}, {``Learning phrase representations using {RNN} encoder-decoder for statistical machine translation,''} in \emph{Proceedings of EMNLP 2014}, 2014, pp. 1724--1734. Available: \url{https://aclanthology.org/D14-1179/}}

\cslhypertarget{ref-friston2017}
\CSLLeftMargin{{[}26{]} }%
\CSLRightInline{K. Friston, T. FitzGerald, F. Rigoli, P. Schwartenbeck, and G. Pezzulo, {``Active inference: A process theory,''} \emph{Neural Computation}, vol. 29, no. 1, pp. 1--49, 2017, doi: \href{https://doi.org/10.1162/NECO_a_00912}{10.1162/NECO\_a\_00912}.}

\cslhypertarget{ref-fountas2020}
\CSLLeftMargin{{[}27{]} }%
\CSLRightInline{Z. Fountas, N. Sajid, P. A. M. Mediano, and K. Friston, {``Deep active inference agents using {Monte-Carlo} methods,''} in \emph{Advances in neural information processing systems 33}, 2020, pp. 11662--11675. Available: \url{https://arxiv.org/abs/2006.04176}}

\cslhypertarget{ref-catal2020}
\CSLLeftMargin{{[}28{]} }%
\CSLRightInline{O. Çatal, S. Wauthier, C. De Boom, T. Verbelen, and B. Dhoedt, {``Learning generative state space models for active inference,''} \emph{Frontiers in Computational Neuroscience}, vol. 14, p. 574372, 2020, doi: \href{https://doi.org/10.3389/fncom.2020.574372}{10.3389/fncom.2020.574372}.}

\cslhypertarget{ref-pinchetti2022}
\CSLLeftMargin{{[}29{]} }%
\CSLRightInline{L. Pinchetti, T. Salvatori, Y. Yordanov, B. Millidge, Y. Song, and T. Lukasiewicz, {``Predictive coding beyond {Gaussian} distributions,''} in \emph{Advances in neural information processing systems 35}, 2022. Available: \url{https://arxiv.org/abs/2211.03481}}

\cslhypertarget{ref-millidge2022b}
\CSLLeftMargin{{[}30{]} }%
\CSLRightInline{B. Millidge, T. Salvatori, Y. Song, R. Bogacz, and T. Lukasiewicz, {``Predictive coding: Towards a future of deep learning beyond backpropagation?''} \emph{arXiv preprint arXiv:2202.09467}, 2022, Available: \url{https://arxiv.org/abs/2202.09467}}

\cslhypertarget{ref-friston2023}
\CSLLeftMargin{{[}31{]} }%
\CSLRightInline{K. Friston \emph{et al.}, {``Path integrals, particular kinds and strange things,''} \emph{Physics of Life Reviews}, vol. 47, pp. 35--62, 2023, doi: \href{https://doi.org/10.1016/j.plrev.2023.08.016}{10.1016/j.plrev.2023.08.016}.}

\cslhypertarget{ref-neftci2019}
\CSLLeftMargin{{[}32{]} }%
\CSLRightInline{E. O. Neftci, H. Mostafa, and F. Zenke, {``Surrogate gradient learning in spiking neural networks: Bringing the power of gradient-based optimization to spiking neural networks,''} \emph{IEEE Signal Processing Magazine}, vol. 36, no. 6, pp. 51--63, 2019, Available: \url{https://arxiv.org/abs/1901.09948}}

\cslhypertarget{ref-bellec2020}
\CSLLeftMargin{{[}33{]} }%
\CSLRightInline{G. Bellec \emph{et al.}, {``A solution to the learning dilemma for recurrent networks of spiking neurons,''} \emph{Nature Communications}, vol. 11, p. 3625, 2020, doi: \href{https://doi.org/10.1038/s41467-020-17236-y}{10.1038/s41467-020-17236-y}.}

\cslhypertarget{ref-zhou2022b}
\CSLLeftMargin{{[}34{]} }%
\CSLRightInline{Z. Zhou \emph{et al.}, {``Spikformer: When spiking neural network meets transformer,''} in \emph{Proceedings of ICLR 2023}, 2022. Available: \url{https://arxiv.org/abs/2209.15425}}

\cslhypertarget{ref-zhu2023}
\CSLLeftMargin{{[}35{]} }%
\CSLRightInline{R.-J. Zhu, Q. Zhao, G. Li, and J. K. Eshraghian, {``{SpikeGPT}: Generative pre-trained language model with spiking neural networks,''} \emph{arXiv preprint arXiv:2302.13939}, 2023, Available: \url{https://arxiv.org/abs/2302.13939}}

\end{CSLReferences}

\end{document}